# LY-GWM-RoboCasa-Human300

## Prediction-Guided Action Selection with a Graph Dynamics Model

TianLiang Ma; Ting Ma; ZhongYuan Ma

Beijing Miwang Information Technology Co., Ltd., Beijing, China

## Abstract

We present LY-GWM-RoboCasa-Human300, a policy that combines a Human300-fine-tuned GR00T-N1.5 action generator, an action-conditioned graph dynamics model, and a learned binary state-reward scorer. At inference, the policy samples four action chunks, predicts the future latent features and robot state for each, and executes the candidate with the highest predicted-state reward logit. Predictions therefore influence the actions sent to the environment. The evaluated system uses a 16-step prediction horizon and keeps all three components frozen during evaluation. In the local evaluation on 50 RoboCasa 1.0.1 tasks in the pretrain kitchen split, with 50 episodes per task, the system records 207/2,500 successes (8.28%), compared with 181/2,500 (7.24%) in the historical self-trained GR00T-N1.5 baseline. The observed difference is 26 successes, or 1.04 percentage points. The atomic group improves from 141/900 to 150/900. Separately, the RoboCasa maintainers re-evaluated the submission and reported 354/1,500 successes (23.6%) across 50 tasks, with leaderboard rates of 48.3%, 16.3%, and 3.1% for Atomic-Seen, Composite-Seen, and Composite-Unseen [9]. The local and official results are reported separately; their difference is not attributed to a model improvement. Recorded local audits identify 212,700 changed environment-input chunks, including 28,223 in atomic tasks. We describe the actual training stages, implemented inference mechanism, evaluation provenance, and public reproduction entrypoints. These results document an active prediction-and-selection system and a positive historical difference.

## 1. Introduction

An action-generating policy can produce several plausible actions for the same observation. Selecting among them requires some assessment of their consequences. Our system separates action proposal, future-state prediction, and candidate scoring: GR00T proposes a small set of actions, LY-GWM predicts candidate-dependent outcomes, and a learned scorer selects the action to execute. This report documents the concrete RoboCasa implementation and its evaluated prediction-and-selection pipeline.

The contribution is an inference-time composition: one shared GR00T backbone evaluation supports four action-head samples, which are explicitly evaluated through a trained latent graph model and reward head. We provide the training specification, exact interfaces, group and task results, action-selection evidence, and limitations needed to assess this composition. The current report concerns the joint-policy evaluation (S42; identifiers are explained in Section 2.1). The released implementation records the active four-candidate prediction-and-selection mechanism [5].

## 2. Background and scope

GR00T supplies the pretrained policy foundation and action-generation architecture [2,3]. RoboCasa365 supplies the household manipulation benchmark and Human300 task setting [1]. The earlier LY-GWM preprint motivates separating structured world representation from action-conditioned transitions [4]. Here, that idea is instantiated as latent graph prediction for candidate selection. The evaluated representation comprises 16 pooled backbone-feature nodes and a robot-state node. It implements action-conditioned latent dynamics and learned predicted-state scoring within the broader design direction of LY-GWM.

PRTS is a public benchmark model with a method paper, code, checkpoints, and evaluation details [8]. Its contrastive reasoning differs from our same-frame reward scoring. We follow conventional method-and-experiment reporting with explicit component attribution.

### 2.1 Component names and experiment identifiers

R5, W5, S38, S31, and S42 are internal workflow identifiers used in the released files. They are not standard model names, parameter counts, or successive versions of a single model. This report pairs each identifier with its functional name so that readers can connect the method to the corresponding configuration and evaluation records.

| Identifier | Functional name | Meaning in this report |
|---|---|---|
| R5 | GR00T-N1.5 policy fine-tuning | Trains the action generator on Human300; produces the step-120,000 policy checkpoint. |
| W5 | LY-GWM graph dynamics training | Trains the model that predicts future features and robot state from a candidate action chunk. |
| S38 | Learned state-reward scorer training | Trains the binary reward MLP used to score the predicted states during joint inference. |
| S31 | Self-trained baseline evaluation | Evaluates the R5 policy alone; records 181/2,500 successes on pretrain. |
| S42 | Prediction-guided joint-policy evaluation | Evaluates the R5 policy, W5 dynamics, and S38 scorer together; records 207/2,500 successes on pretrain. |
| S30 | Public-checkpoint reference evaluation | Records a separate local evaluation of the public GR00T-N1.5 checkpoint on target. |
| S32 | Historical shadow-prediction evaluation | Computes graph predictions alongside the policy; the current active-selection results refer to S42. |
| W3 / W4 / S37 | Training-data preparation stages | W3 selects transition pairs; W4 caches frozen-policy features; S37 prepares recorded reward labels. |

**Training and evaluation are distinct.** Policy fine-tuning (R5), graph dynamics training (W5), and reward scorer training (S38) produce three separately trained components. Baseline evaluation (S31) uses the fine-tuned action policy alone. Joint-policy evaluation (S42) composes the same policy architecture with the trained dynamics and scorer; all three components remain frozen while actions are selected and executed. The labels provide provenance across files, while the functional names describe what the system does.

## 2.2 Related work and positioning

Generative action policies. Diffusion Policy [10] models visuomotor actions through conditional denoising and generates action sequences for receding-horizon execution. GR00T [2,3] supplies the vision-language-action foundation used in this report. These works motivate separating the ability to propose plausible actions from the decision about which proposal to execute. Our evaluated generator is the Human300-fine-tuned GR00T-N1.5 policy, not a newly trained Diffusion Policy. At each request, four flow-matching action samples share one backbone evaluation; a separately trained dynamics model and scorer then rank them.

Prediction-based planning and world-model learning. Visual Foresight [11] learns visual predictions from robot interaction and uses them for model-predictive control. TD-MPC2 [12] performs trajectory optimization in the latent space of a learned world model. Both place prediction within action decision-making. Our implementation is a restricted candidate-selection procedure: it predicts one endpoint at $t + 16$ for each of four proposed chunks and selects the highest state-reward logit. It does not perform iterative trajectory optimization, use a terminal value function, or claim the planning capabilities of those systems.

Learning behavior through imagination. DreamerV3 [13] learns an environment model and improves behavior through imagined futures. DayDreamer [14] applies world-model learning directly to physical robots, including manipulation and locomotion. The connection is the use of learned action consequences to support control. Our three components are trained separately from recorded data and remain frozen during evaluation. The evaluated system performs neither online reinforcement learning nor actor-critic updates on imagined trajectories.

Graph-structured dynamics. Interaction Networks [15] model object interactions through relational computations on graphs and demonstrate learned physical dynamics. Our predictor also uses message passing, but its 17 nodes consist of 16 pooled backbone-feature nodes and one robot-state node. The feature nodes are not supervised object identities, and the learned edges are not established physical or causal relations. This distinction limits any interpretation of the present graph as an object-level physical simulator.

Recent world-model-to-action approaches. RoboActualizer [16], by Du, Xie, and colleagues, uses a frozen pretrained world-model encoder and a lightweight actualizer with interacting latent and action branches, jointly predicting future latents and actions by flow matching. It addresses the related problem of turning predictive representations into executable behavior. Our system instead retains a distinct action generator, learns an action-conditioned graph predictor, and explicitly scores the predicted endpoints of proposed actions. The training objectives and inference mechanisms differ; this report provides no matched experimental comparison with RoboActualizer.

Scope of the contribution. Prediction-guided control, generative action policies, and graph dynamics are established research directions. Our contribution is the documented composition and evaluation of these components in the RoboCasa Human300 setting. The reported local historical difference and independent official re-evaluation do not establish superiority over the related methods or unrestricted task generalization.

# 3. Implemented method

## 3.1 Inputs and candidate generation

Let o denote the camera observations, s the normalized 20-dimensional robot state, and l the task instruction. The frozen GR00T backbone processes the policy input. Valid backbone tokens are adaptively average-pooled to 16 feature nodes of width 2,048 and normalized per node, yielding F. The policy's padded state interface is B x 1 x 64; its padded action output is B x 16 x 32. The graph consumes the first 20 state components and the 16 x 12 effective normalized action components [5].

Candidate zero is the original action-head output. Three additional head calls use the same cached backbone output and separately seeded random streams. The cache is captured before the action head can mutate its input. On the first request, restored random state and cached-head replay are checked against the original output, with maximum absolute difference at most 1e-6. The backbone is not run four times. Each candidate uses four flow-matching integration steps. Additional-candidate seeding is deterministic from the recorded task, request context, and candidate identity.

## 3.2 Action-conditioned graph dynamics

Sixteen feature nodes and one state node are projected to width 384 and augmented with learned node embeddings. A two-layer MLP projects the flattened 192-dimensional action chunk into a width-384 control vector u. Four graph blocks use directed, fully connected message passing with self-messages masked out. For N = 17 nodes, block updates are:

```
u = action_MLP(vec(A))
m_i = (1 / 16) sum_{j != i} message_MLP([h_i, h_j, u])
h_i_next = LayerNorm(h_i + update_MLP([h_i, m_i]))      (1)
```

The message MLP has dimensions 1,152 -> 768 -> 384; the update MLP has dimensions 768 -> 768 -> 384, both with SiLU. Separate output heads predict residuals, which are added to the input feature nodes and state. The resulting dynamics model D has 10,237,204 parameters and predicts the state at t + 16 from the current representation and a complete candidate chunk.

## 3.3 Predicted-state scoring and selection

For each candidate k, the predicted feature nodes are mean-pooled and concatenated with the predicted state, giving 2,068 inputs. Stored training-set means and scales normalize these inputs. A 2,068 -> 128 -> 1 SiLU MLP produces a reward logit:

```
(F_hat_k, s_hat_k) = D(F, s, A_k)
z_k = reward_MLP(normalize([mean_nodes(F_hat_k), s_hat_k]))
k_star = argmax_k z_k,  k in {0, 1, 2, 3}             (2)
```

Exact ties select the earliest candidate. The scorer has 264,961 trainable parameters and no explicit goal input. Task information reaches candidate generation through GR00T; the scorer operates on its predicted latent features and state. It is trained on recorded same-frame binary reward, not discounted return or a calibrated probability of eventual rollout success.

**Algorithm 1. One policy request.** Encode current observations once; retain candidate zero; sample three additional action chunks; predict four future feature/state pairs; score all four predictions; select the maximum logit; pass the selected padded output through GR00T's original inverse transforms; execute up to 16 actions; observe again and repeat. All model weights are frozen. Environment rewards and observed-future diagnostic probes do not enter selection. Medoid computations in the source are diagnostic; the evaluated decision mode is learned_reward.

## 4. Training and data

Training comprises three separate stages. The graph and reward stages do not update GR00T. No model is updated during S42 evaluation. Public configurations and model implementations are pinned in [5,6]; the released resources are described in Section 8.

| Stage | Data and supervision | Trained components | Schedule |
|---|---|---|---|
| R5 policy | Human300; 30,000 demonstrations; expert actions | GR00T projector and action head | 120,000 updates; global batch 128 |
| W5 dynamics | 108,000 train / 12,000 validation transition pairs | Graph dynamics only | 10 epochs; 33,750 updates; batch 32 |
| S38 reward | 216,000 train / 24,000 validation observed states | Reward MLP; separate state-only diagnostic head | 5 epochs; batch 512 |

### 4.1 GR00T-N1.5 policy fine-tuning (R5)

R5 starts from nvidia/GR00T-N1.5-3B at revision 869830fc749c35f34771aa5209f923ac57e4564e, using the RoboCasa GR00T source revision 9d7d7a9eb7ad30bd8ce30448d9ab53a918b45b10. The selected dataset is pretrain_human300, split pretrain, filter 100_demos: 100 episodes for each of 300 tasks. Data configuration is panda_omron with embodiment new_embodiment. The language and visual backbone parameters are frozen; the projector and action model are fine-tuned.

Training uses 16 GPUs, per-device batch 2, gradient accumulation 4, AdamW with learning rate 3e-5, weight decay 1e-5, betas (0.95, 0.999), epsilon 1e-8, gradient norm cap 1.0, cosine schedule, and 6,000 warmup steps. BF16 and TF32 are enabled in this training stage. The final checkpoint is step 120,000. Global batch times updates equals 15,360,000 sample presentations, not that many distinct demonstrations.

The inherited flow-matching objective interpolates expert action A and Gaussian noise e: $X_t = (1-t)e + tA$. The head predicts velocity A-e with masked mean-squared error over valid action dimensions. Inference starts from noise and uses four Euler updates with step size 1/4. The implementation's configuration name tune_diffusion_model refers to this action model; it should not be interpreted as a separate noise-prediction or DDIM objective.

### 4.2 LY-GWM graph dynamics training (W5)

For downstream model fitting, each task's 100 episodes are divided into 90 training and 10 validation episodes. Four 16-step transitions per episode yield 108,000 training and 12,000 validation pairs. For episode length L, sampling indices follow floor(i(L-17)/3), i = 0,1,2,3, with the future at t+16. Frozen R5 features, 20D states, and demonstrated 16 x 12 actions define the prediction targets. This split holds out data from W5 and S38, not from R5, which used all 30,000 demonstrations.

W5 minimizes mean feature MSE plus mean state MSE with equal coefficients. It starts from fresh parameters with zero-initialized residual output layers. It uses AdamW (weight decay 1e-4, gradient cap 1.0), cosine learning rate 1e-4 to 1e-6, 200 warmup steps, seed 4892026, FP32 with TF32 disabled, maximum 10 epochs, and patience 3. Training completes all 33,750 updates; epoch 10 is selected by validation loss.

### 4.3 State-reward scorer training (S38)

S38 uses current and future observed states from the cached pairs, labeled by the corresponding recorded binary reward. Inputs are normalized by training mean and population standard deviation with a 0.05 scale floor. A separate 20 -> 64 -> 1 state-only head is trained as an offline diagnostic. The two BCE losses are summed; their combined parameters use AdamW, learning rate 3e-4, weight decay 1e-4, gradient cap 1.0, and seed 4892038. Output biases start at the training-prior logit. Checkpoint selection uses observed-validation BCE. Predictions from W5 on 12,000 validation demonstration-action pairs are diagnostic only; they do not train or select the scorer. These details were checked against the archived S38 worker whose SHA-256 matches the public configuration (Appendix B).

# 5. Local evaluation protocol and results

## 5.1 Benchmark setting

The evaluated S42 system uses RoboCasa 1.0.1, pretrain kitchens, 18 Atomic-Seen tasks, 16 Composite-Seen tasks, and 16 Composite-Unseen tasks. Each task has 50 episodes, giving group denominators 900, 800, and 800. The official benchmark uses the 50 target task identities in pretraining kitchens; the phrase "target tasks" does not imply the environment split named target [1,7]. The exact per-task horizons are listed in Appendix A.

Five environment slots are processed per task. Constructor seed is 489000 + 1000 x task_index + episode_index. The policy process uses seed 489 + task_index. The action chunk length is 16 and the action head uses four integration steps. Success is the worker's logical OR of chunk-end info.success over the episode. Episodes run under the recorded task horizon and completion logic; no successful-only or bound-clean subset is substituted for the full denominator.

Original execution used NVIDIA A100 40GB GPUs, Linux/Python 3.10, PyTorch 2.5.1+cu121 and EGL rendering. Policy dependencies include NumPy 1.26.4 and Transformers 4.51.3; the separate simulation environment uses NumPy 2.2.5 and RoboCasa 1.0.1. Inference disables TF32. The release preserves observed environment records, not a universally tested installation lock [5].

## 5.2 Comparison with the self-trained baseline

| Group | Historical S31 baseline | S42 joint policy | Success delta | Rate delta (pp) |
|---|---|---|---|---|
| Atomic-Seen | 141/900 (15.6667%) | 150/900 (16.6667%) | +9 | +1.000 |
| Composite-Seen | 25/800 (3.125%) | 38/800 (4.750%) | +13 | +1.625 |
| Composite-Unseen | 15/800 (1.875%) | 19/800 (2.375%) | +4 | +0.500 |
| Overall | 181/2,500 (7.24%) | 207/2,500 (8.28%) | +26 | +1.040 |

**Table 1.** All 2,500 episodes are included in each arm. Overall is the episode-weighted rate, equivalent here to averaging the 50 task rates: (18 r_atomic + 16 r_seen + 16 r_unseen)/50. "pp" denotes percentage points. Higher group aggregates do not imply improvement on every task.

The S31 and S42 manifests and recorded checkpoint paths align. Their exported episode identities also match in task, constructor seed, horizon, and chunk length. This comparison uses the historical checkpoint paths and recorded metadata for provenance and reports the observed outcomes under those recorded settings.

## 5.3 Offline dynamics validation

| Error on 12,000 held-out W5 pairs | Persistence | W5 prediction |
|---|---|---|
| Feature MSE | 0.02689410 | 0.01713613 |
| State MSE | 0.01001973 | 0.00071013 |
| Sum of the two means | 0.03691383 | 0.01784625 |

**Table 2.** Persistence copies the current feature/state to the future. Validation is used for W5 checkpoint selection and contains demonstrated actions. The two error measures quantify future-feature and robot-state prediction on this offline demonstration distribution.

## 5.4 Official independent re-evaluation

Following the local experiments documented in Sections 5.1-5.3, the RoboCasa maintainers independently re-evaluated the submitted system. The maintainer published the reproduced task rates and aggregate counts on 28 September 2026 in leaderboard PR #19. The submission was merged on 30 September 2026 and appears on the official leaderboard as LY-GWM-RoboCasa, submitted by SEIN-LYGWM [9].

The official report covers the same 50 task identities, grouped into 18 Atomic-Seen, 16 Composite-Seen, and 16 Composite-Unseen tasks. It reports 540, 480, and 480 episodes for the three groups, respectively. These totals and the reported task rates correspond to 30 episodes per task, totaling 1,500 episodes. This differs from the 50 episodes per task used in the local evaluations.

| Group | Tasks | Successes / episodes | Leaderboard rate |
|---|---|---|---|
| Atomic-Seen | 18 | 261 / 540 | 48.3% |
| Composite-Seen | 16 | 78 / 480 | 16.3% |
| Composite-Unseen | 16 | 15 / 480 | 3.1% |
| Overall | 50 | 354 / 1,500 | 23.6% |

Table 1a. Official RoboCasa re-evaluation [9]. Group percentages reproduce the one-decimal official leaderboard display. Overall is calculated from the exact counts: (261 + 78 + 15) / (540 + 480 + 480) = 354/1,500 = 23.6%.

### 5.4.1 Relationship to the local experiments

The local S31 baseline (181/2,500; 7.24%) and S42 joint policy (207/2,500; 8.28%) remain the historical comparison reported in Table 1. The +1.04 percentage-point difference, paired outcome analysis, bootstrap interval, and action-input diagnostics apply only to those local records. The S30 public-checkpoint reference (212/2,500; 8.48%) remains a separate local evaluation on the target kitchen split, as described in Section 7.

The official 23.6% is an independent re-evaluation result, not a replacement denominator for the local experiment and not a new local ablation. Its difference from 8.28% must not be interpreted as an improvement caused by a model update or as an isolated contribution of the graph dynamics model. The public re-evaluation comment does not specify every runtime setting; the archived local seeds, horizons, dependency versions, and diagnostic counts are therefore not asserted to describe the official run.

### 5.4.2 Official task-level results and source

Appendix A.1 reproduces all 50 task success rates from the maintainer's comment. Appendix A retains the original local counts, policy comparison, and recorded horizons. The official source and leaderboard are cited in [9].

# 6. Outcome and action-selection analyses

## 6.1 Exploratory paired outcome accounting

We rejoin the public S31 outcome table with S42 episode records using task and episode indices, checking the recorded seed, horizon, task group, and chunk length. This analysis was added for this report on 25 September 2026; it does not represent a new rollout or a prespecified experimental test.

| Group | Both succeed | S31 only | S42 only | Both fail |
|---|---|---|---|---|
| Atomic-Seen | 93 | 48 | 57 | 702 |
| Composite-Seen | 7 | 18 | 31 | 744 |
| Composite-Unseen | 5 | 10 | 14 | 771 |
| Overall | 105 | 76 | 102 | 2,217 |

**Table 3.** The 26-success aggregate difference equals 102 gains minus 76 losses. At task level, 20 task counts increase, 13 decrease, and 17 remain unchanged (Appendix A). Matching constructor seeds identifies records; it does not ensure identical later states after policies take different actions.

A two-sided exact binomial calculation on the 178 discordant pairs gives p = 0.06065 under the independent-discordant-pair null. This is a descriptive calculation: task clustering and the historical design limit its interpretation. A separate exploratory percentile bootstrap resamples task-level success-count differences within each group, retaining 18/16/16 tasks, with 20,000 replicates and NumPy default_rng seed 20260925. Its 95% interval is [0.00, 2.12] percentage points. This resamples the task mix, not repeated policy seeds. These calculations summarize uncertainty in the recorded historical comparison. The supplement provides the exact calculation script and derived counts.

## 6.2 Evidence that selection affects environment inputs

| Group | Action chunks | Non-first selected | Changed input chunks |
|---|---|---|---|
| Atomic-Seen | 37,550 | 28,223 | 28,223 |
| Composite-Seen | 116,100 | 86,862 | 86,862 |
| Composite-Unseen | 130,150 | 97,615 | 97,615 |
| Overall | 283,800 | 212,700 | 212,700 |

**Table 4.** Changed inputs are measured relative to candidate zero in the same policy request. These are chunks passed to the environment, not post-controller motor commands. The recorded audits also mark prediction and score divergence on all 283,800 chunks. The selector is active in atomic and composite tasks; task-group membership does not disable it.

The non-first selection fraction is 74.947%. Together, the source and selection/execution reports document active candidate selection. Action-chunk counts describe mechanism usage; task success is measured separately at episode level in Tables 1 and 3.

## 6.3 Result aggregation and action-input diagnostics

The reported totals use S42_ACCEPTANCE_V2_BOUNDS_DIAGNOSTIC_ONLY. This collection rule treats action-bound counters as diagnostics and retains episode, report, and evidence checks. It was applied after the rollout, replacing the original zero-count acceptance condition without changing recorded actions, seeds, rewards, or outcomes. The original collector and revision diff remain available.

The records contain 116,968 out-of-bound scalar elements in selected action chunks and 116,292 in consumed input prefixes, across 1,822 episodes and all 50 tasks, including 149 successful episodes. These are environment-input measurements. The worker applies the original action inverse transforms without extra clipping or denormalization, and the reported score includes all 2,500 episodes.

# 7. Public-checkpoint reference results

## 7.1 Evaluation setting and recorded results

The historical S30 local evaluation of the public GR00T-N1.5 checkpoint records 212/2,500 successes (8.48%). The archived manifest explicitly sets split = target. In contrast, S31 and S42 use split = pretrain, and the official multi-task leaderboard evaluates in pretraining kitchens [7]. The public-checkpoint result is therefore reported as a separate reference, with its split explicitly identified.

| Historical S30, target split | Atomic-Seen | Composite-Seen | Composite-Unseen | Overall |
|---|---|---|---|---|
| Successes / episodes | 162/900 | 36/800 | 14/800 | 212/2,500 |
| Success rate | 18.00% | 4.50% | 1.75% | 8.48% |

**Table 5.** Separate historical reference, not a same-protocol baseline for S42. Task-group labels describe task membership; they do not remove the kitchen-split difference. The supplied author audit identifies the S30 protocol digest in Appendix B.

As checked on 25 September 2026, the official leaderboard displays GR00T N1.5 at 23.9% overall (50.7%, 14.8%, 2.7% by group), and notes reevaluation with a 1.5x longer horizon relative to the paper's results [7].

# 8. Public artifacts and reproduction resources

The public release provides inference implementations, reviewer entrypoints, GR00T upstream training source, stage configurations, policy and dynamics training summaries, inference weights, episode records, per-task audit summaries, and recorded trace hashes [5,6]. These materials connect the method description to fixed implementation and result records.

The training documentation covers policy fine-tuning (R5), graph dynamics training (W5), and reward scorer training (S38). The released execution entrypoints focus on inference and evaluation; full retraining additionally uses custom data-preparation and training orchestration, caches, and reward labels. A new evaluation supplies compatible local environments and simulator assets as described in Section 9. Historical verification operates on exported episode records and audit summaries, while a fresh evaluation generates its own complete selection/execution traces.

## 9. Reproduction procedure

Use the fixed code and model revisions below, rather than assuming that a mutable main branch reproduces the evaluated system. Full URLs and source mappings appear in Appendices B-C.

| Component | Immutable revision |
| --- | --- |
| Evaluation code | 79b247c854df116614d9ec5db29406c329a18603 |
| Hugging Face inference release | 180883d9af9ee02edc3c53fe20babdc7658d1127 |
| Upstream GR00T training source | 9d7d7a9eb7ad30bd8ce30448d9ab53a918b45b10 |
| NVIDIA base-model revision | 869830fc749c35f34771aa5209f923ac57e4564e |

**Step 1: verify exported records.** In the complete pinned repository, run:

```
python3 -B reviewer_s42/verify_historical_results.py
```

For this report, the check was executed on the downloaded fixed code snapshot. It verified exported file hashes, all 2,500 episode records, 207 successes, receipt digests, and recorded action-evidence totals. This is a CPU record-integrity check of the exported results. New GPU evaluation follows the preparation and execution procedure below.

**Step 2: prepare a new run.** Follow reviewer_s42/README.md. Supply existing policy and simulation environments, the fixed model snapshot, and RoboCasa/robosuite/robomimic sources and simulator assets. The following paths are explicit placeholders for the reviewer's installation:

```
python3 -B reviewer_s42/prepare_run.py \
  --weights /absolute/weights_snapshot \
  --policy-env /absolute/policy_env \
  --sim-env /absolute/simulation_env \
  --third-party /absolute/third_party \
  --output /absolute/new_s42_review_run
```

Preparation checks required source, configuration, and weight hashes. It relocates path assignments while checking that the worker's remaining syntax tree and protocol content are unchanged. The new output directory must be outside the code repository and must not already exist. Relocation produces a new protocol digest while preserving the original digest. No historical episode scores are copied.

**Step 3: evaluate every task on allocated GPUs, then collect.** The first command below runs task 0 only; complete indices 0 through 49 for the full evaluation. The scripts do not allocate or submit GPU jobs.

```
python3 -B reviewer_s42/run_task.py \
  --run /absolute/new_s42_review_run --task-index 0
python3 -B reviewer_s42/collect_results.py \
  --run /absolute/new_s42_review_run
```

The new-run collector checks complete coverage and the newly generated selection/execution traces. Partial-task continuation and mixing episodes from different jobs are unsupported. Do not disable Python assertions with -O. The archived cluster job scripts contain historical paths and are not the reviewer entrypoint. A fresh run may obtain a different success count.

## 10. Conclusion

LY-GWM-RoboCasa-Human300 implements explicit prediction-guided action selection on top of a Human300-fine-tuned GR00T-N1.5 policy. In S42, predictions are used to choose executed actions, including in atomic tasks. The local recorded result is 207/2,500 successes, with a positive historical difference of 26 successes over S31. Separately, official re-evaluation records 354/1,500 successes (23.6%) [9]. The report makes the actual training and inference design inspectable, identifies the public-checkpoint evaluation split, and connects the results to reproducible records. The released checkpoints and reviewer entrypoints support independent benchmark evaluation of this system.

## Appendix A. Local task results and horizons

Every row contains 50 episodes per policy. H is the recorded maximum environment-step horizon. Counts are successes, not percentages. S31 denotes baseline-policy evaluation and S42 joint-policy evaluation; delta is joint minus baseline. Indices are those used by the seed formula and reviewer entrypoints. A = Atomic-Seen; C = Composite-Seen; U = Composite-Unseen.

| Id | Task | Group | H | S31 | S42 | Delta |
|---|---|---|---|---|---|---|
| 0 | CloseBlenderLid | A | 900 | 1 | 1 | +0 |
| 1 | CloseFridge | A | 900 | 10 | 11 | +1 |
| 2 | CloseToasterOvenDoor | A | 450 | 4 | 6 | +2 |
| 3 | CoffeeSetupMug | A | 600 | 1 | 1 | +0 |
| 4 | NavigateKitchen | A | 450 | 3 | 2 | -1 |
| 5 | OpenCabinet | A | 1050 | 12 | 17 | +5 |
| 6 | OpenDrawer | A | 750 | 10 | 10 | +0 |
| 7 | OpenStandMixerHead | A | 450 | 11 | 16 | +5 |
| 8 | PickPlaceCounterToCabinet | A | 750 | 11 | 14 | +3 |
| 9 | PickPlaceCounterToStove | A | 600 | 16 | 11 | -5 |
| 10 | PickPlaceDrawerToCounter | A | 750 | 8 | 11 | +3 |
| 11 | PickPlaceSinkToCounter | A | 900 | 10 | 10 | +0 |
| 12 | PickPlaceToasterToCounter | A | 600 | 15 | 13 | -2 |
| 13 | SlideDishwasherRack | A | 450 | 7 | 8 | +1 |
| 14 | TurnOffStove | A | 750 | 3 | 0 | -3 |
| 15 | TurnOnElectricKettle | A | 450 | 6 | 9 | +3 |
| 16 | TurnOnMicrowave | A | 450 | 3 | 2 | -1 |
| 17 | TurnOnSinkFaucet | A | 600 | 10 | 8 | -2 |
| 18 | DeliverStraw | C | 2550 | 0 | 0 | +0 |
| 19 | GetToastedBread | C | 3000 | 0 | 0 | +0 |
| 20 | KettleBoiling | C | 1500 | 1 | 3 | +2 |
| 21 | LoadDishwasher | C | 1800 | 3 | 1 | -2 |
| 22 | PackIdenticalLunches | C | 3900 | 0 | 5 | +5 |
| 23 | PreSoakPan | C | 2400 | 3 | 5 | +2 |
| 24 | PrepareCoffee | C | 1800 | 0 | 0 | +0 |
| 25 | RinseSinkBasin | C | 1350 | 3 | 6 | +3 |
| 26 | ScrubCuttingBoard | C | 1200 | 4 | 5 | +1 |
| 27 | SearingMeat | C | 4350 | 1 | 0 | -1 |
| 28 | SetUpCuttingStation | C | 2400 | 2 | 1 | -1 |
| 29 | StackBowlsCabinet | C | 2100 | 1 | 2 | +1 |
| 30 | SteamInMicrowave | C | 2100 | 0 | 0 | +0 |
| 31 | StirVegetables | C | 2400 | 0 | 3 | +3 |
| 32 | StoreLeftoversInBowl | C | 2550 | 6 | 5 | -1 |
| 33 | WashLettuce | C | 1650 | 1 | 2 | +1 |
| 34 | ArrangeBreadBasket | U | 4350 | 1 | 0 | -1 |
| 35 | ArrangeTea | U | 2250 | 1 | 0 | -1 |
| 36 | BreadSelection | U | 1950 | 2 | 5 | +3 |
| 37 | CategorizeCondiments | U | 1650 | 0 | 1 | +1 |
| 38 | CuttingToolSelection | U | 1200 | 0 | 0 | +0 |
| 39 | GarnishPancake | U | 2700 | 4 | 3 | -1 |
| 40 | GatherTableware | U | 2250 | 0 | 0 | +0 |
| 41 | HeatKebabSandwich | U | 2700 | 0 | 0 | +0 |
| 42 | MakeIceLemonade | U | 3000 | 0 | 0 | +0 |
| 43 | PanTransfer | U | 1800 | 0 | 0 | +0 |
| 44 | PortionHotDogs | U | 2250 | 0 | 2 | +2 |
| 45 | RecycleBottlesByType | U | 2850 | 4 | 4 | +0 |
| 46 | SeparateFreezerRack | U | 2400 | 0 | 0 | +0 |
| 47 | WaffleReheat | U | 4050 | 1 | 2 | +1 |
| 48 | WashFruitColander | U | 3150 | 2 | 2 | +0 |
| 49 | WeighIngredients | U | 3000 | 0 | 0 | +0 |
| Total | 50 tasks x 50 episodes | A/C/U | - | 181 | 207 | +26 |

# Appendix A.1. Official re-evaluation task results

Rates below are transcribed from the RoboCasa maintainer's re-evaluation [9]. The group totals correspond to 30 episodes per task. A = Atomic-Seen; C = Composite-Seen; U = Composite-Unseen. Indices match the task identities in the local appendix; local horizons and success counts are not substituted for official measurements.

| Id | Task | Group | Episodes | Official rate |
|---|---|---|---|---|
| 0 | CloseBlenderLid | A | 30 | 13.33% |
| 1 | CloseFridge | A | 30 | 53.33% |
| 2 | CloseToasterOvenDoor | A | 30 | 36.67% |
| 3 | CoffeeSetupMug | A | 30 | 16.67% |
| 4 | NavigateKitchen | A | 30 | 26.67% |
| 5 | OpenCabinet | A | 30 | 73.33% |
| 6 | OpenDrawer | A | 30 | 73.33% |
| 7 | OpenStandMixerHead | A | 30 | 80.00% |
| 8 | PickPlaceCounterToCabinet | A | 30 | 70.00% |
| 9 | PickPlaceCounterToStove | A | 30 | 66.67% |
| 10 | PickPlaceDrawerToCounter | A | 30 | 60.00% |
| 11 | PickPlaceSinkToCounter | A | 30 | 96.67% |
| 12 | PickPlaceToasterToCounter | A | 30 | 36.67% |
| 13 | SlideDishwasherRack | A | 30 | 56.67% |
| 14 | TurnOffStove | A | 30 | 13.33% |
| 15 | TurnOnElectricKettle | A | 30 | 23.33% |
| 16 | TurnOnMicrowave | A | 30 | 40.00% |
| 17 | TurnOnSinkFaucet | A | 30 | 33.33% |
| 18 | DeliverStraw | C | 30 | 3.33% |
| 19 | GetToastedBread | C | 30 | 0.00% |
| 20 | KettleBoiling | C | 30 | 23.33% |
| 21 | LoadDishwasher | C | 30 | 23.33% |
| 22 | PackIdenticalLunches | C | 30 | 10.00% |
| 23 | PreSoakPan | C | 30 | 40.00% |
| 24 | PrepareCoffee | C | 30 | 0.00% |
| 25 | RinseSinkBasin | C | 30 | 30.00% |
| 26 | ScrubCuttingBoard | C | 30 | 46.67% |
| 27 | SearingMeat | C | 30 | 6.67% |
| 28 | SetUpCuttingStation | C | 30 | 13.33% |
| 29 | StackBowlsCabinet | C | 30 | 3.33% |
| 30 | SteamInMicrowave | C | 30 | 6.67% |
| 31 | StirVegetables | C | 30 | 10.00% |
| 32 | StoreLeftoversInBowl | C | 30 | 13.33% |
| 33 | WashLettuce | C | 30 | 30.00% |
| 34 | ArrangeBreadBasket | U | 30 | 0.00% |
| 35 | ArrangeTea | U | 30 | 0.00% |
| 36 | BreadSelection | U | 30 | 6.67% |
| 37 | CategorizeCondiments | U | 30 | 0.00% |
| 38 | CuttingToolSelection | U | 30 | 0.00% |
| 39 | GarnishPancake | U | 30 | 13.33% |
| 40 | GatherTableware | U | 30 | 0.00% |
| 41 | HeatKebabSandwich | U | 30 | 0.00% |
| 42 | MakeIceLemonade | U | 30 | 3.33% |
| 43 | PanTransfer | U | 30 | 0.00% |
| 44 | PortionHotDogs | U | 30 | 0.00% |
| 45 | RecycleBottlesByType | U | 30 | 10.00% |
| 46 | SeparateFreezerRack | U | 30 | 0.00% |
| 47 | WaffleReheat | U | 30 | 6.67% |
| 48 | WashFruitColander | U | 30 | 3.33% |
| 49 | WeighIngredients | U | 30 | 6.67% |
| Total | 50 tasks; 354 successes | A/C/U | 1,500 | 23.6% |

## Appendix B. Provenance and source map

The supplement contains per-task CSV, paired counts, bootstrap settings, the record-verification result, a sanitized S30 provenance summary, and a script that reproduces the added analysis from the pinned repository. Digests below preserve their original meanings: file-byte SHA-256 and canonical JSON-object SHA-256 are not interchangeable.

| Object / digest kind | SHA-256 |
|---|---|
| S42 protocol / canonical JSON | 4bdd5abfabe2674caba78c1f7b637f3c141d5550a09b55e065020c8a9c069e71 |
| S31 protocol / recorded digest | ee15b8a67247a3f37f6e137c222e0b79cc1801c6d0ab539ec6db247e32bbc943 |
| S30 protocol / recorded digest | 7af171be0a1bafad91e6ffe5a90fe2032a89f8e1428ce4f9f6e59cb27f9bed2e |
| R5 weight shard 1 / file bytes | a734b821ddf1a145536baeeb5b144e0e8435f1746ee6aced523d2ff0a19d5d4a |
| R5 weight shard 2 / file bytes | 0b713131a2e4627710a57fdf65887f44629d505e15ec66df18ac32a47c7485d3 |
| W5 weights / file bytes | a964d19a0078cc6282d5da79f06f7d7ead3137799b8f6e969ccff5e0643be9f1 |
| S38 weights / file bytes | 5f9ad4b52eddfc2ca3f1ac06efcae3f20a02a764857987fe447810d5cde68a8b |
| W5 archived worker / file bytes | aa6bb5eece2eca77b2a23625a138336ef02132852f24e9498882ffa1fb4d1a2c |
| S38 archived worker / file bytes | 52ef967efff2747bb11e9a0d34fa8a76a212203d87bcdb431ff04dfbecbe78b3 |

| Claim | Source within the pinned evaluation repository |
|---|---|
| R5 data, fine-tuning, schedule | evaluated_snapshot/gr00t_training/formal_config.json; formal_results.json |
| W5 architecture and validation | evaluated_snapshot/w5_training/graph_model.py; config.json; training_results.json |
| Scorer inputs, target, provenance | evaluated_snapshot/s42/s42_original/scorer-training-config.json; full/learned_reward/reward_model.py |
| Candidate sampling and execution | evaluated_snapshot/s42/s42_original/full/learned_reward/decision_adapter.py; selection_rule.py; worker.py |
| Protocol and per-task horizons | evaluated_snapshot/s42/s42_original/full/learned_reward/protocol.json |
| Revised result and receipt | evaluated_snapshot/s42/acceptance_v2/overall_results.json; per_task_results.json; evaluation_receipt.json |
| Historical baseline episodes | evaluated_snapshot/joint_s32/baseline_episode_outcomes.json |
| New-run and record verification | reviewer_s42/README.md; verify_historical_results.py; prepare_run.py; run_task.py; collect_results.py |

**Author archival evidence.** The W5 and S38 training workers were recovered from the author's installers and byte-hashed. Both match their public configuration digests. This corroborates the optimizer and schedule in Section 4.2 and the normalization and BCE implementation in Section 4.3. The pinned inference repository does not contain these full training workers. The S30 split and result were recovered from the author's read-only evaluation audit dated 23 September 2026. These archival sources are distinguished from publicly downloaded files; the supplement exposes the relevant non-private facts and their recorded digests.

**Component attribution.** NVIDIA supplies the base GR00T model; the RoboCasa team supplies benchmark assets and the GR00T adaptation. We trained the R5 fine-tuned policy, W5 dynamics model, and S38 scorer. All three are frozen during S42. Access and reuse follow the notices and component terms preserved in the respective repositories; this report does not replace those terms.

## Appendix C. References

## Data and code availability

The fixed code and weights are available at [5,6]. The reviewer guide is reviewer_s42/README.md in [5]. The statistical supplement distributed with this report adds analysis of existing exported outcomes; it adds no new GPU experiment. The conceptual preprint [4] and this implementation report serve different purposes and should be linked together when describing the submission.

## Appendix C. References (continued)